%% file: main_arxiv.tex
\documentclass[10pt,twocolumn,letterpaper]{article}

\usepackage{iccv}    

\input{math_commands.tex}

\usepackage{url}
\usepackage{comment}
\usepackage{amsmath,amssymb} 
\usepackage{multirow}
\usepackage{booktabs}
\usepackage{subcaption}
\usepackage{bm}
\usepackage{array}
\usepackage{xcolor}
\usepackage{graphicx}
\usepackage{enumitem}
\usepackage{pgfplots}
\usepackage[accsupp]{axessibility}
\usepackage{cuted}

\usepackage{wrapfig}

\usepackage[pagebackref=true,breaklinks=true,letterpaper=true,colorlinks,bookmarks=false]{hyperref}

\usepackage[capitalize]{cleveref}
\crefname{section}{Sec.}{Secs.}
\Crefname{section}{Section}{Sections}
\Crefname{table}{Table}{Tables}
\crefname{table}{Tab.}{Tabs.}

\iccvfinalcopy
\def\iccvPaperID{9343} 

\ificcvfinal\fi

\newcommand{\modelname}{PT$^2$}

\begin{document}

\title{Collecting The Puzzle Pieces: Disentangled Self-Driven Human Pose Transfer by Permuting Textures}

\author{Nannan Li\\
Boston University\\
{\tt\small nnli@bu.edu}
\and
Kevin J Shih\\
NVIDIA\\
\and
Bryan A. Plummer \\
Boston University\\
{\tt\small bplum@bu.edu}
}
\maketitle
\ificcvfinal\thispagestyle{empty}\fi

\begin{abstract}
Human pose transfer synthesizes new view(s) of a person for a given pose. Recent work achieves this via self-reconstruction, which disentangles a person's pose and texture information by breaking the person down into parts, then recombines them for reconstruction. However, part-level disentanglement preserves some pose information that can create unwanted artifacts. In this paper, we propose Pose Transfer by Permuting Textures (\modelname{}), an approach for self-driven human pose transfer that disentangles pose from texture at the patch-level. Specifically, we remove pose from an input image by permuting image patches so only texture information remains. Then we reconstruct the input image by sampling from the permuted textures for patch-level disentanglement. To reduce noise and recover clothing shape information from the permuted patches, we employ encoders with multiple kernel sizes in a triple branch network.
On DeepFashion and Market-1501, \modelname{} reports significant gains on automatic metrics over other self-driven methods, and even outperforms some fully-supervised methods. A user study also reports images generated by our method are preferred in 68\% of cases over self-driven approaches from prior work. Code is available at \url{https://github.com/NannanLi999/pt_square}
\end{abstract}

\section{Introduction}
\begin{figure}[!t]
\centering
\begin{tabular}{c|c}
    \centering
    \begin{subfigure}[t]{0.235\textwidth}
  \centering
    \includegraphics[width=\textwidth]{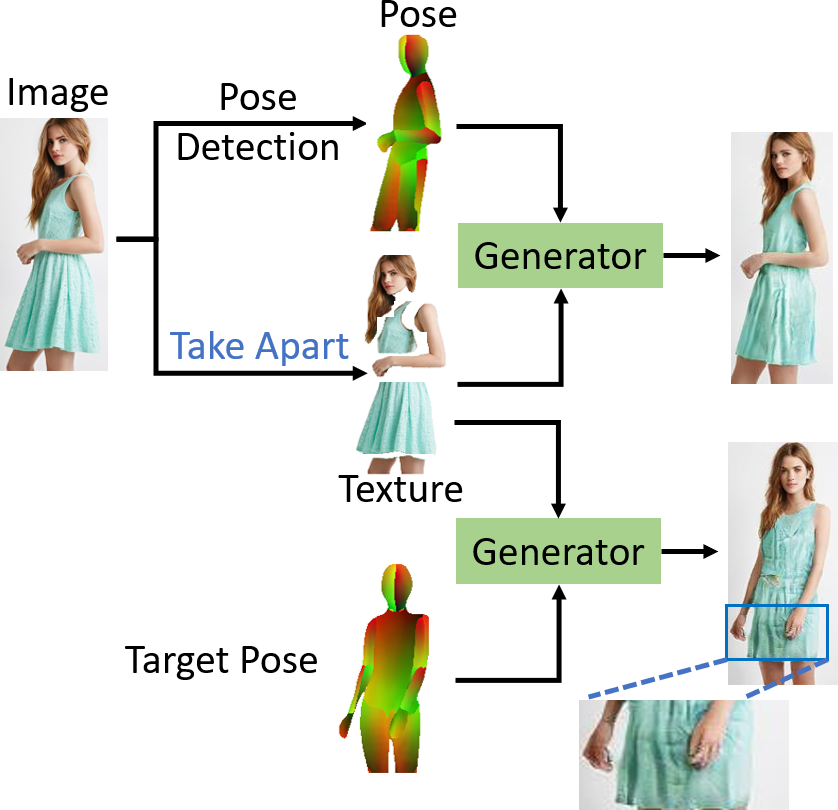}
    \caption{Prior work.}
    \label{fig:motiv_prior}
 \end{subfigure}
 &
  \begin{subfigure}[t]{0.235\textwidth}
  \centering
    \includegraphics[width=\textwidth]{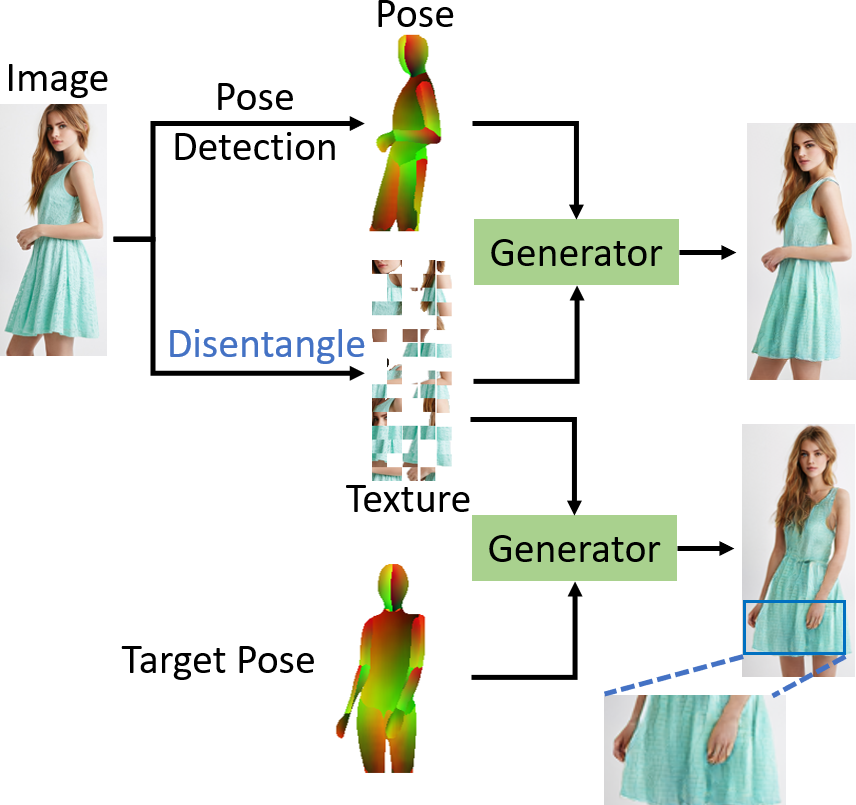}
    \caption{Our approach.}   
    \label{fig:bin}
 \end{subfigure}
  \end{tabular}
  \vspace{-2mm}
    \caption{Self-driven pose transfer methods extract texture and pose representations and then learn to reconstruct the original image.
    \textbf{(a)} Most recent work breaks down the person into several parts to disentangle textures \cite{ma2018disentangled,ma2021must,wang2022self}. However, pose information may still appear in the part-wise texture features. Without supervision, disentangling them is difficult \cite{locatello2019challenging}. 
    \textbf{(b)} Our approach disentangles textures from pose by permuting the image patches, effectively eliminating pose information, which enables more effective separation of pose and texture features than prior work.}
 \label{fig:motiv}
 \vspace{-4mm}
\end{figure}

The goal of human pose transfer is to change the pose of a person while preserving the person's appearance and clothing textures. It has wide applications such as virtual try-on \cite{yang2020towards-unsup,cui2021dressing,yu2019vtnfp}, controllable person image manipulation \cite{cui2021dressing,liu2021liquid} and person re-identification \cite{zhang2021seeing}. Recent work focused on using paired image data (\ie, two images of the same person before and after reposing) \cite{zhou2022cross,zhang2022exploring}, but collecting such data can be very labor intensive. Alternatively, self-driven methods train pose transfer models without paired images \cite{ma2021must,song2019unsupervised}. However, they face two major challenges: effectively disentangling texture and pose, and preserving texture details across changes in pose. As illustrated in Fig.\ \ref{fig:motiv_prior}, most prior works attempt to achieve the pose and texture disentanglement at a part-level \cite{ma2018disentangled, yang2020towards,ma2021must,wang2022self}, but each body part may retain some pose information. Without direct supervision from pose-invariant textures, disentangling pose and textures in the image is difficult \cite{locatello2019challenging}. 

To address the above issues, we propose Pose Transfer by Permuting Textures (\modelname{}), a self-driven pose transfer method that uses input permutation to transfer clothing patterns to the target pose. The input permutation function geometrically disentangles texture from pose in raw images without requiring supervision. As shown in Fig.\ \ref{fig:bin}, this creates a disentangled texture sample space by randomly reordering the texture patches on the person so that the source pose is not easily recovered from the permuted textures. We find this patch-level disentanglement can transfer detailed clothing patterns to any target pose. However, our permutation function does introduce two new issues that we must address to effectively transfer pose.  


The first drawback we address is that in addition to losing pose information after permutation, we also lose shape information. To recover some shape information of clothing items (\eg, short dress or long dress), we use a dense pose representation \cite{guler2018densepose} to model the geometric transformation between pose representations.  While prior work also leverages pose information, this is typically a sparse representation (\eg,~\cite{ma2021must,wang2022self}).  As our experiments show, \modelname{} more effectively leverages the denser representations than prior work. 

The second issue we address arises from the unnatural boundaries between newly neighboring permuted patches. Specifically, we use a dual-scale encoder in the pose and texture branches for handling permuted inputs. The small-scale encoder provides a more accurate representation of the texture details without the permutation noise, whereas the large-scale encoder enables us to capture longer range visual patterns. Our experiments show that each encoder provides a unique and necessary representation of the texture patterns, resulting in best performance when they are used jointly. 


Our paper is organized as follows. Sec.\ \ref{sec:method} introduces our model pipeline. Sec.\ \ref{sec:ip} details our paper's main contribution, the input permutation function. Sec.\ \ref{sec:dke} describes the dual-scale encoder for recovering the clothing shape from the permuted textures. Extensive experiments in Sec.\ \ref{sec:exp} show that \modelname{} significantly improves the image quality of self-driven approaches on DeepFashion \cite{liu2016deepfashion} and Market-1501 \cite{zheng2015scalable}. The user study in Sec.\ \ref{sec:quan} also reports that our synthesized images are preferred in 68\% over prior work.

\section{Related Work}

\noindent\textbf{Pose transfer with unpaired images.} Transferring pose without losing texture details when paired data is absent is challenging \cite{wang2022self}. 
Early attempts produced poor quality images due to overfitting to an identity mapping in the self-supervised training \cite{pumarola2018unsupervised,esser2018variational,ma2018disentangled}. 
More recent methods use cycle-GAN to include both source and target poses in training \cite{song2019unsupervised,sanyal2021learning}, or disentangle pose and textures to prevent overfitting \cite{albahar2021pose,wang2022self,fu2022stylegan,ma2021must}. These include using the StyleGAN framework to learn implicitly disentangled features \cite{albahar2021pose,fu2022stylegan}, or incorporating a part-wise texture encoder \cite{wang2022self,lorenz2019unsupervised}. 
However, we find that input permuting is more effective at disentangling texture and pose than part-wise methods.  This remains true even when incorporating methods that erase pose information in part-wise features by using their mean and variance  as style vectors (\eg, as done in~\cite{ma2021must}). As we will show, such transformations results in washed-out clothing details in the style representation. 
In contrast to these methods, our approach disentangles the texture at a patch level, and then restores the clothing details using dual-scale encoders in a triple branch network, which achieves better texture transfer with large pose variations.
\smallskip

\noindent\textbf{Pose transfer with paired images.} 
Methods trained with paired images learn the clothing deformation via soft attention that aggregates source person features with weighted sampling \cite{zhu2019progressive,zhang2022exploring,ren2022neural,gao2020recapture}, or via sparse attention that approximates a dense flow field from the source to the target person \cite{han2019clothflow,tang2021structure,ren2020deep}. Some methods also use a semantic parsing map to provide further guidance to control the style of each body part \cite{zhang2021pise,lv2021learning}. To help infer occluded parts of the person after pose transfer, prior work has also explored inpainting 2D partial texture to 3D full texture in the UV space, and then projecting it back to the 2D pose \cite{grigorev2019coordinate,sarkar2020neural,albahar2021pose,sarkar2021style}. However, the 2D to 3D projection could lose partial texture information. Another line of work \cite{xu2022surface,shao2022doublefield} uses multiple views of the same person to achieve 3D reconstruction with various poses. These methods all require full supervision from paired data, which might be difficult to collect in some real-world scenarios. Our approach only needs unpaired data for training, which enables direct fitting to an in-the-wild target domain.
\smallskip

\noindent\textbf{Jigsaw Puzzle Solving.} Our approach is also similar to some self-supervised representation learning methods that jigsaw puzzle solving, where the pretext task divides the image into large patches and then infers their relative positions by using each patch's inherent geometry information  (\eg,~\cite{noroozi2016unsupervised,carlucci2019domain}). However, our goal is to \emph{sample} relevant patches based on the target posture. Thus, with the target posture as guidance, we use much smaller image patches to remove position information and to disentangle pose and texture.

\begin{figure*}[!t]
    \centering
    \includegraphics[width=\textwidth]{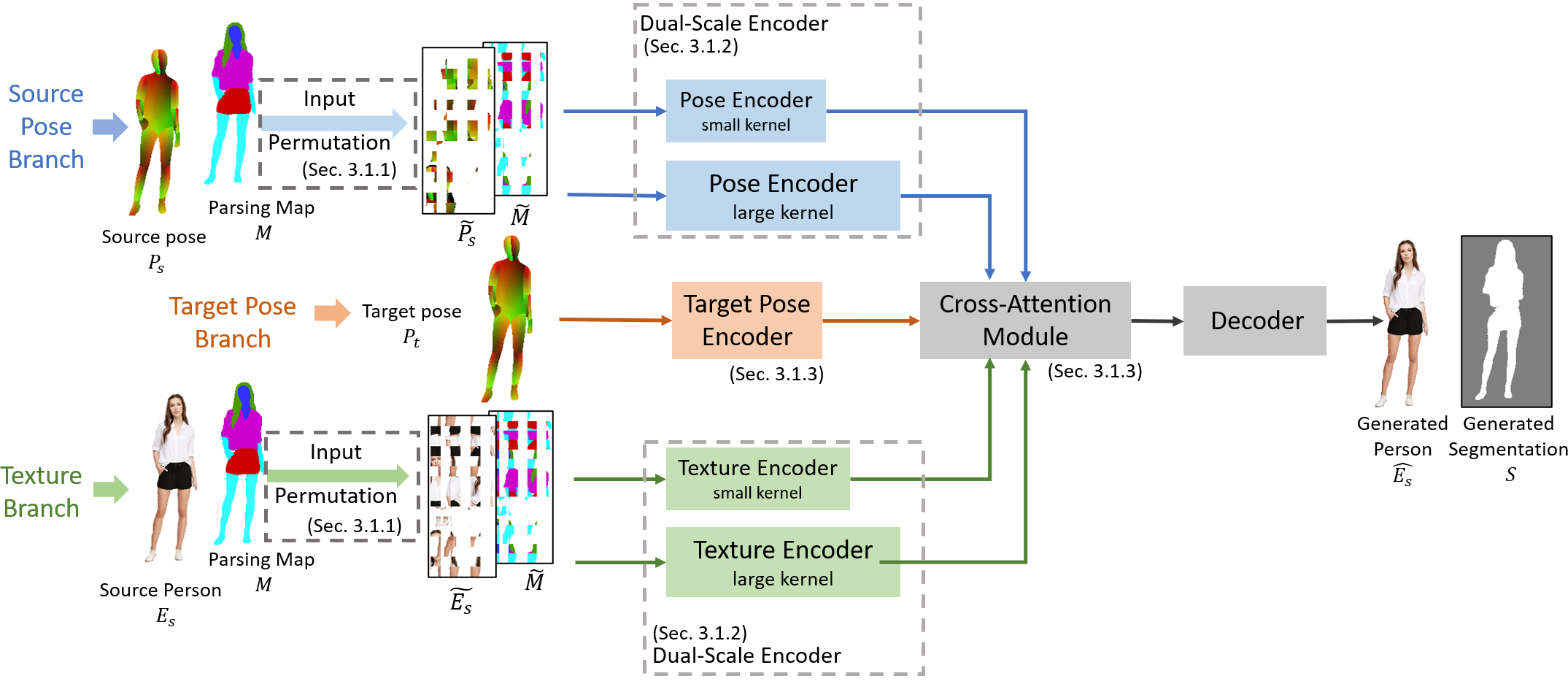}
    \vspace{-4mm}
    \caption{Overview of our pose transfer network in \modelname{}.  The network takes the source person $E_s$, source parsing map $M$, and source pose $P_s$ as inputs. In the source pose branch and texture branch, $P_s$ and $E_s$ are first permuted (Sec.\ \ref{sec:ip}) to create the corresponding sample space, which is encoded with dual-scale encoders (Sec.\ \ref{sec:dke}). Then the encoded features are sampled in a cross-attention module (Sec.\ \ref{sec:att}) to be decoded into the generated person $\hat{E}_s$ and its segmentation $S$. The output of the pose transfer network is combined with the output of the background inpainting network (Sec.\ \ref{sec:bin}) to produce the final image.}
    \vspace{-4mm}
 \label{fig:model}
\end{figure*}

\section{Self-Driven Pose Transfer by Permuting Textures}
\label{sec:method}
 Let $I_s$ be the source image with posture $P_s$. Our goal is to synthesize a new view $I_t$ of the same person wearing the same clothes in a target posture $P_t$. We assume we only have \emph{unpaired} data for training, which enables direct fitting to an in-the-wild target domain as our experiments will show. 
Following~\cite{dundar2021unsupervised,liu2021liquid}, we separate the image $I_s$ into a foreground person $E_s$ and the background $B$. The foreground person is generated by our pose transfer network (shown in Fig.\ \ref{fig:model} and discussed Sec.\ \ref{sec:ptn}). The background is produced by the background inpainting network in Sec.\ \ref{sec:bin}. They are combined to create the final reconstructed image $\hat{I}_s$. We will describe our training objectives in Sec.\ \ref{sec:loss}.

\subsection{Pose Transfer Network}
\label{sec:ptn}
The goal of the pose transfer network is to generate a new view of the foreground person in a target pose. As illustrated in Fig.\ \ref{fig:model}, we use three branches: a source pose branch, a target pose branch, and a texture branch. 
The target pose branch encodes the given target pose. 
The source pose and texture branch first randomly permutes the source pose and texture, respectively (Sec.\ \ref{sec:ip}), then uses a dual-scale encoder (Sec.\ \ref{sec:dke}) to encode the permuted patches. The features of the permuted patches are sampled via cross-attention based on the target pose features (Sec.\ \ref{sec:att}). The resulting outputs in the source pose branch and texture branch are then merged and decoded into the generated person $\hat{E}_s$ and its segmentation $S$. The rest of the image, namely the background for $\hat{E}_s$ and $S$, will be generated from a specialized background inpainting network (Sec.\ \ref{sec:bin}) to create the final image. As shown in Fig.\ \ref{fig:model}, the target pose is identical to the source pose during training. However, during inference we simply need to replace the target pose with a new (different) posture to enable pose transfer.

\subsubsection{Input Permutation}
\label{sec:ip}
The input permutation in Fig.\ \ref{fig:model} disentangles the pose and textures at patch-level by randomly shuffling image patches, which is introduced below.
\smallskip

\noindent\textbf{Inputs to the Texture Branch.} As illustrated for the texture branch in Fig.\ \ref{fig:model}, the texture representation should not simply be the source person $E_s$ itself, as it is entangled with its posture. To erase the pose information from $E_s$, we create a texture sample space by dividing the image into $p\times p$ squares, referred to as ``patches,'' then shuffling their locations. Intuitively the original pose cannot easily be retrieved from the permuted patches when the patch size $p$ is sufficiently small. Additionally, we mask 20\% of the patches so the model can learn to reconstruct occluded regions. Formally, let $\text{RandMask}(\cdot)$ be the input permutation function. The inputs of the texture branch become $[\tilde{E_s};\tilde{M}]=\text{RandMask}([E_s;M],m_t)$, where $[;]$ is concatenation and $m_t$ is the masking rate. We set $m_t=0.2$ in our experiments (see Sec.\ \ref{sec:abl} for an ablation study).  The permuted textures $[\tilde{E_s};\tilde{M}]$ are given as inputs to the dual-scale texture encoder (Sec.\ \ref{sec:dke}).
\smallskip
\begin{figure}[t]
  \centering
    \includegraphics[width=0.35\textwidth]{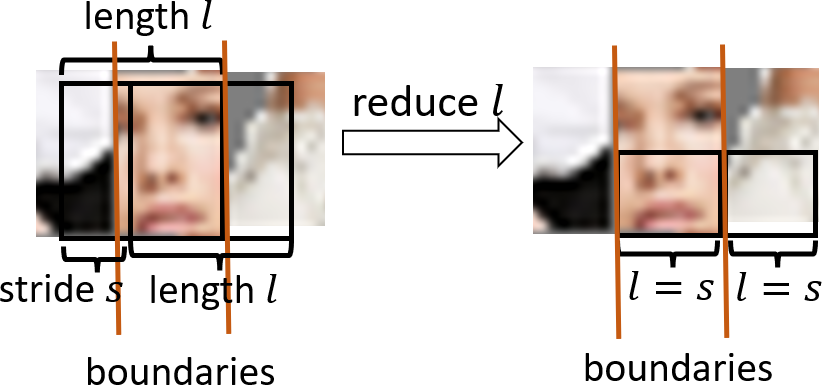}
 
    \caption{Comparing the receptive field in the large-kernel encoder (left) and small-kernel encoder (right) on a closeup of two permuted patches. The kernel size $l$ is reduced to avoid boundary-crossing (see Sec.\ \ref{sec:dke}).
    }
    \label{fig:rpf} 
\end{figure}

\noindent\textbf{Inputs to the Pose Branches.} Prior work uses a single branch to encode pose and texture \cite{pumarola2018unsupervised,ma2018disentangled,ma2021must,wang2022self}, but we find separate source pose and texture branches (as shown in Fig.\ \ref{fig:model}) improves shape and texture recovery after pose transfer. These pose branches do not contain texture information as postures are represented as 2D UV coordinates obtained from DensePose \cite{guler2018densepose} pretrained on COCO \cite{lin2014microsoft}.  We also permute source pose the same way as textures to learn a pose transformation function that supports large pose variations. In addition, we mask 50\% of the source pose features to force the model to learn the inherited symmetry in the human body. Thus, inputs to the source pose branch are $[\tilde{P_s};\tilde{M}]=\text{RandMask}([P_s;M],m_p)$, where $m_p=0.5$. During training, the target pose is the original pose, but it is replaced with a new pose during inference. Note that the inputs of the texture and source pose branches are permuted the same way so they are spatially aligned in the attention module (see Sec.\ \ref{sec:att}). The permuted source pose representations $[\tilde{P_s};\tilde{M}]$ are then fed to a dual-scale pose encoder (Sec.\ \ref{sec:dke}).%

\subsubsection{Dual-Scale Encoder}
\label{sec:dke}
We encode the permuted source pose/texture inputs from Sec.\ \ref{sec:ip} with a dual-scale encoder containing twelve convolutional layers. When compared to the single-scale encoder in prior work \cite{pumarola2018unsupervised,ma2018disentangled,ma2021must,wang2022self}, we use different kernel sizes in the encoder's convolutional layers to address issues that arise due to the boundaries in permuted patches. Specifically, each feature vector has a certain receptive field in the image. For a receptive field that crosses the boundary between two permuted image patches, the pixels within the field could be spatially distant and irrelevant to each other due to the permutation. For example, in the left picture of Fig.\ \ref{fig:rpf}, the black squares denote two adjacent receptive fields. Both squares crossing the boundary are seeing two distinct patterns (\eg, face and shoes) within their own receptive fields. This could introduce a high volume of noise to the feature vector, preventing the model from recognizing true clothing patterns.

A simpler solution to avoid boundary-crossing is to encode each patch separately. However, this would limit the effective receptive field to be at most the patch size and thus prevent the model from learning more styles and patterns. To solve the issue, we combine the large-kernel encoder with an additional encoder with a reduced kernel size such that the length of the receptive field $l$ equals its stride $s$ and is a divisor of the patch size $p$. We select this small kernel size such that the kernel does not cross the boundary of the permuted inputs from Sec.\ \ref{sec:ip}. As shown in the right picture of Fig.\ \ref{fig:rpf}, this design avoids the boundary-crossing problem, enabling the convolutional kernel to learn a consistent pattern within its receptive field. Note that a large kernel size is still necessary as it has more parameters and a larger receptive field for learning larger image patterns. By combining encoders with large and small kernels, our model is capable of learning better feature representations from permuted patches. 
\begin{figure}[t]
  \centering
    \includegraphics[width=0.5\textwidth]{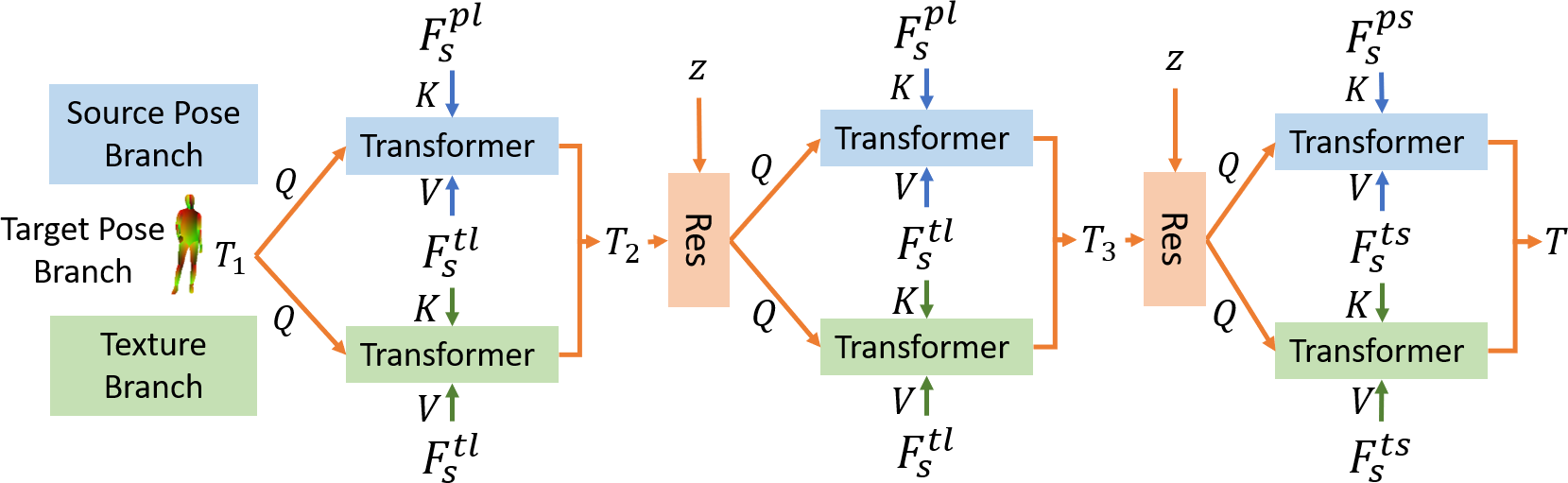}
    \caption{Cross-attention module in Sec.\ \ref{sec:att}. Res represents a residual layer. The cross-attention mechanism aligns the permuted texture with the target pose.}  
    \label{fig:att}
\end{figure}
\subsubsection{Cross-Attention Module}
\label{sec:att}
The cross-attention module takes the output feature map of the dual-kernel encoders (Sec.\ \ref{sec:dke}) as inputs and learns to sample texture based on the target pose. As shown in Fig.\ \ref{fig:att}, the attention module consists of three vision transformers \cite{tang2020xinggan,zhang2022exploring} formulated as $\text{Attention}(Q,K,V)=\text{softmax}(\frac{QK^T}{\sqrt{d}})\cdot V$. 
\smallskip

\noindent\textbf{Sampling Source Pose Features}. The first two transformers of the source pose branch sample permuted pose features based on the attention between the poses. Let $F_s^{pl}, F_s^{tl}$ represent the output feature map of the large-kernel source pose encoder and texture encoder, respectively. $T_1$ is the feature map of the target pose $P_t$, which is encoded by the target pose encoder. The attention is learned from the correlation between the source pose feature $F_s^{pl}$ and the target person feature ${T_i}$, which is computed using the transformer's attention function as:
\begin{equation}
     Q=W_i^{pq}{T_i}, K=W_i^{pk}F_s^{pl}, V=W_i^{pv}F_s^{tl}, i=1,2.
\end{equation}

\noindent Here, $W_i^{pq},W_i^{pk},W_i^{pv}$ are learnable projection matrices. After each transformer layer, the output features are concatenated and passed through a residual layer \cite{he2016deep} as in Fig.\ \ref{fig:att}. A random noise vector $z$ is injected into the residual layer as the affine transformation parameters of the feature map $T_i$ to prevent mode collapse \cite{karras2019style}. The residual layer can be written as: $\text{Res}(T_i)=f_i(z)\cdot {h_i}(T_{i})+g_i(z)+{h_i^{skip}}(T_{i})$, where $f_i,g_i,h_i,h_i^{skip}$ are learnable functions.

Note that the first transformer learns a texture-agnostic attention map for the target pose. Starting from $T_2$, information from the texture branch is gradually integrated into the target person representation $T_2$ and $T_3$. 
\smallskip

\noindent \textbf{Sampling Textures.} The texture branch keeps refining the sampled textures based on the target person representation $T_i$. In order to produce spatially aligned feature maps in the sample space, this branch uses the same architectural design as the source pose branch. The attention map for texture sampling is formulated as the correlation between the texture feature $F_s^{tl}$ and the target person feature $T_i$:
\begin{equation}
    Q=W_i^{tq}{T_i}, K=W_i^{tk}F_s^{tl}, V=W_i^{tv}F_s^{tl}, i=1,2.
\end{equation}

\noindent \textbf{Fusing features from the small kernel encoder.}  Let $F_s^{ps}, F_s^{ts}$ represent the output feature map of the small-kernel source pose encoder and texture encoder, respectively. In the last transformer layer, we replace $F_s^{pl},F_s^{tl}$ in the above equations with features produced by small-kernel encoders (\ie, $F_s^{ps},F_s^{ts}$) to learn more fine-grained image details. The output feature map $T$ of the last transformer layer is then fed to the decoder, where $T$ is gradually upsampled to the target person $\hat{E}_s$ and its segmentation mask $S$.

The cross-attention between two pose branches learns a geometric transformation between different postures. Then the texture branch samples the given textures based on the target pose. Fusing the two sources of information provides a more accurate match between the given pose and textures. By filling in more clothing details that are learned through small kernels, our pose transfer network can then faithfully recover the appearance of clothing items after pose transfer.

\subsection{Background Inpainting Network}
\label{sec:bin}
Following \cite{dundar2021unsupervised,liu2021liquid}, we use a separate UNet \cite{long2015fully}-based background inpainting network to infer the background pixels of the masked foreground region. During training, we mask the bounding box containing the source person so the model learns to infer the entire background area during pose transfer. We combine the generated person with the output of the background inpainting network to obtain the final image. Formally, let $\hat{B}$ be the inpainted background output. Given the generated person $\hat{E}_s$ and its segmentation mask $S$ produced by the pose transfer network, the final reconstructed image is: 
\begin{equation}
    \hat{I}_s=S \odot \hat{E}_s + (1-S) \odot \hat{B}
\end{equation}

\subsection{Training Objectives}
\label{sec:loss}
We train \modelname{} using an adversarial loss using a pair of discriminators $D$ and $D_p$ \cite{ma2021must}. $D$ penalizes the distribution difference between the synthesized image $\hat{I}_s$ and the ground truth $I_s$. $D_p$ evaluates that if the posture in $\hat{I}_s$ matches the source pose $P_s$. Thus, our adversarial loss is written as:
\begin{equation}
\begin{aligned}
    L_{adv}= &D(\hat{I}_s)^2 + (1-D(I_{s}))^2 \\
    &+D_p([\hat{I}_s;P_s])^2+(1-D_p([I_{s};P_s]))^2.
\end{aligned}
\end{equation}

To ensure the correctness of our image generation, we first add a simple reconstruction loss,
\begin{equation}
    L_{rec}=||\hat{I}_s-I_{s}||_1.
\end{equation}
We also include a perceptual loss \cite{johnson2016perceptual} that encourages both the ground truth and reconstructed image to have similar semantic properties,
\begin{equation}
    L_{perc}=\sum\nolimits_i ||{{\phi ^i}({\hat{I}_s}) - {\phi ^i}({I_s})}||_1,
\end{equation}
where ${\phi ^i}$ is the $i$th layer of a VGG model \cite{simonyan2014very} pretrained on ImageNet~\cite{deng2009imagenet}.  Finally, we use a style loss that penalizes discrepancies on colors and textures using the Gram matrix $\mathbb{G}(\cdot)$ of the features,
\begin{equation}
    L_{style}=\sum\nolimits_i ||{\mathbb{G}({\phi ^i}({\hat{I}_s})) - \mathbb{G}({\phi ^i}({I_s}))}||_1.
\end{equation}
Thus, our total loss can be written as
\begin{equation}
    L_{total} =\lambda_1 L_{adv}+\lambda_2 L_{rec}+\lambda_3 L_{perc}+\lambda_4 L_{style}.
\end{equation}
where $\lambda_{1-4}$ are scalar hyperparameters. 

\section{Experiments}
\label{sec:exp}
\noindent\textbf{Datasets.} We evaluate our proposed model on two benchmarks: DeepFashion \cite{liu2016deepfashion} and Market1501 \cite{zheng2015scalable}. DeepFashion contains 52,712 high-quality images with a clean background. Market-1501 has 32,668 low-resolution images with various lighting conditions and noisy background. Following \cite{zhang2022exploring,wang2022self}, we select 8,570 test pairs on DeepFashion and 12,000 test pairs on Market-1501. As in prior self-driven methods \cite{ma2021must,wang2022self}, we use 37,332 training images for DeepFashion and 12,112 training images for Market-1501.
\smallskip

\begin{table}[!t]
    \centering
    \setlength{\tabcolsep}{3pt}
    \begin{tabular}{lcccc}
    \toprule
    Method &FID$\downarrow$ &SSIM$\uparrow$ &LPIPS$\downarrow$ &IS$\uparrow$ \cr \midrule    
    \multicolumn{3}{l}{
    \textbf{Supervised by paired images}} 
    \cr
    GFLA$^*$\cite{ren2020deep} &10.513 &0.798 &0.146 &3.246 \cr
    PISE \cite{zhang2021pise} &13.623 &0.777 &0.189 &3.257\cr
    SPIG  \cite{lv2021learning} &12.243 &0.790 &0.186 &3.097 \cr
    DPTN \cite{zhang2022exploring} &13.466 &0.789 &0.176 &3.253 \cr
    CASD$^*$\cite{zhou2022cross} &11.360 &\textbf{0.813} &0.154 &3.056\cr
    NTED \cite{ren2022neural} &\textbf{6.786} &0.808 &\textbf{0.133} &\textbf{3.264} \cr
    NTED(Market)  &37.481 &0.712 &0.307 &3.065\cr
    \midrule
    \textbf{No paired images} \cr
    DPIG \cite{ma2018disentangled} &44.615 &0.705 &0.275 &3.176  \cr
    VU-Net \cite{esser2018variational} &23.012 &0.789 &0.312 &3.012\cr
    E2E \cite{song2019unsupervised} &16.350 &0.761 &0.206 &3.431 \cr
    MUST-GAN \cite{ma2021must} &15.012 &0.773 &0.190 &3.470 \cr
    MUST-GAN(DP) &12.157 &0.781 &0.183 &3.422 \cr
    SCM-Net \cite{wang2022self} &12.180 &0.751 &0.182 &\textbf{3.632} \cr
    \modelname(Ours) &\textbf{8.338} &\textbf{0.795} &\textbf{0.158} &3.469 \cr
    \bottomrule
    \end{tabular}
    \caption{Pose transfer results on DeepFashion. (Market) means the method is trained on Market-1501 and (DP) means the method uses DensePose features. Results from prior work are evaluated using our code except SCM-Net. Generated images of methods marked with $*$ are resized due to different training image sizes.}
    \label{tab:deepfashion}
    \vspace{-3mm}
\end{table} 

\noindent\textbf{Metrics.} Following \cite{song2019unsupervised,zhang2022exploring}, we train our \modelname{} model on 256$\times$176 images and pad the generated images to 256$\times$256 for evaluation in DeepFashion. In Market-1501, we use the same image size (\emph{i.e.}, 128$\times$64) for training and evaluation. We use Structural Similarity Index Measure (SSIM) \cite{wang2004image}, Frechet Inception Distance (FID) \cite{heusel2017gans}, Learned Perceptual Image Patch Similarity (LPIPS) \cite{zhang2018unreasonable}, and Inception Score (IS) \cite{salimans2016improved} to evaluate image synthesis quality. IS assesses the quality of images generated by adversarial training. In Market-1501, we add Masked-SSIM and Masked-LPIPS computed on the target person region to exclude the irrelevant background from contributing to the metrics. 

All the methods are compared using our code under the same image sizes and padding method, except SCM-Net \cite{wang2022self}, which does not have publicly available code. 
Specifically, following \cite{song2019unsupervised}, we pad the original/generated images of size 256 $\times$ 176 to 256 $\times$ 256 for our evaluation in DeepFashion. We pad all the generated images from other methods \cite{ren2020deep,zhang2021pise,lv2021learning,zhang2022exploring,zhou2022cross,song2019unsupervised,ma2021must} in the same manner, and report their scores under our evaluation code in Tables \ref{tab:deepfashion} \& \ref{tab:market}. Note that due to a different training image size in \cite{ren2020deep,zhou2022cross}, we resized their generated images from 256 $\times$ 256 to 256 $\times$ 176 using PIL bilinear resampling and then added the padding. In Market-1501, following \cite{lv2021learning}, we keep the size of the original image (\emph{i.e.}, 128 $\times$ 64), for all comparisons.
More training details are included in Supplementary Sec.\ B.

\begin{table}[!t]
    \centering	
    \setlength{\tabcolsep}{3pt}
    \begin{tabular}{lcccc} 
    \toprule
    Method &FID$\downarrow$ &M-SSIM$\uparrow$  &M-LPIPS$\downarrow$ &IS$\uparrow$ \cr \midrule 
    \multicolumn{5}{l}{
    \textbf{Supervised by paired images}}
    \cr 
    GFLA \cite{ren2020deep} &20.194 &0.815 &0.138 &2.546 \cr  
    SPIG \cite{lv2021learning} &22.043 &0.819 &0.129 &2.761 \cr
    DPTN \cite{zhang2022exploring} &17.929 &\textbf{0.820} &0.125 &2.479 \cr
    NTED(DF) &38.831 &0.734 &0.212 &2.242 \cr
    \midrule
    \multicolumn{5}{l}{
    \textbf{No paired images}} \cr   
    \modelname(Ours) &\textbf{17.389} &\textbf{0.820} &\textbf{0.122} &\textbf{2.789}\cr
    \bottomrule
    \end{tabular}
    \caption{Pose transfer results on Market-1501. Results from prior work are evaluated using our code. (DF) means the method is trained on DeepFashion.} 
    \label{tab:market}
    \vspace{-4mm}
\end{table}

\subsection{Quantitative Results}
\label{sec:quan}
Table \ref{tab:deepfashion} compares methods on the pose transfer task using DeepFashion, where our approach significantly outperforms prior unpaired approaches on FID, SSIM, and LPIPS. Notably, we found our FID to be 4 points better than that of SCM-Net. Our method also outperforms most supervised (paired) methods \emph{without paired training data}. Similar behavior is seen on Market-1501 (Table \ref{tab:market}), where \modelname{} outperforms paired methods in Masked-SSIM and Masked-LPIPS. 

We perform an additional test to rule out dense pose as the primary contributing factor. For this, we retrained MUST-GAN \cite{ma2021must}, replacing their OpenPose with DensePose (denoted as MUST-GAN(DP) in Table \ref{tab:deepfashion}). MUST-GAN(DP) improves the original OpenPose variant, suggesting dense pose as a better representation. However, our proposed approach still outperforms MUST-GAN(DP).

\smallskip

\begin{figure*}[!t]
    \centering
    \begin{subfigure}[c]{0.6\textwidth}
  \centering
    \includegraphics[width=\textwidth]{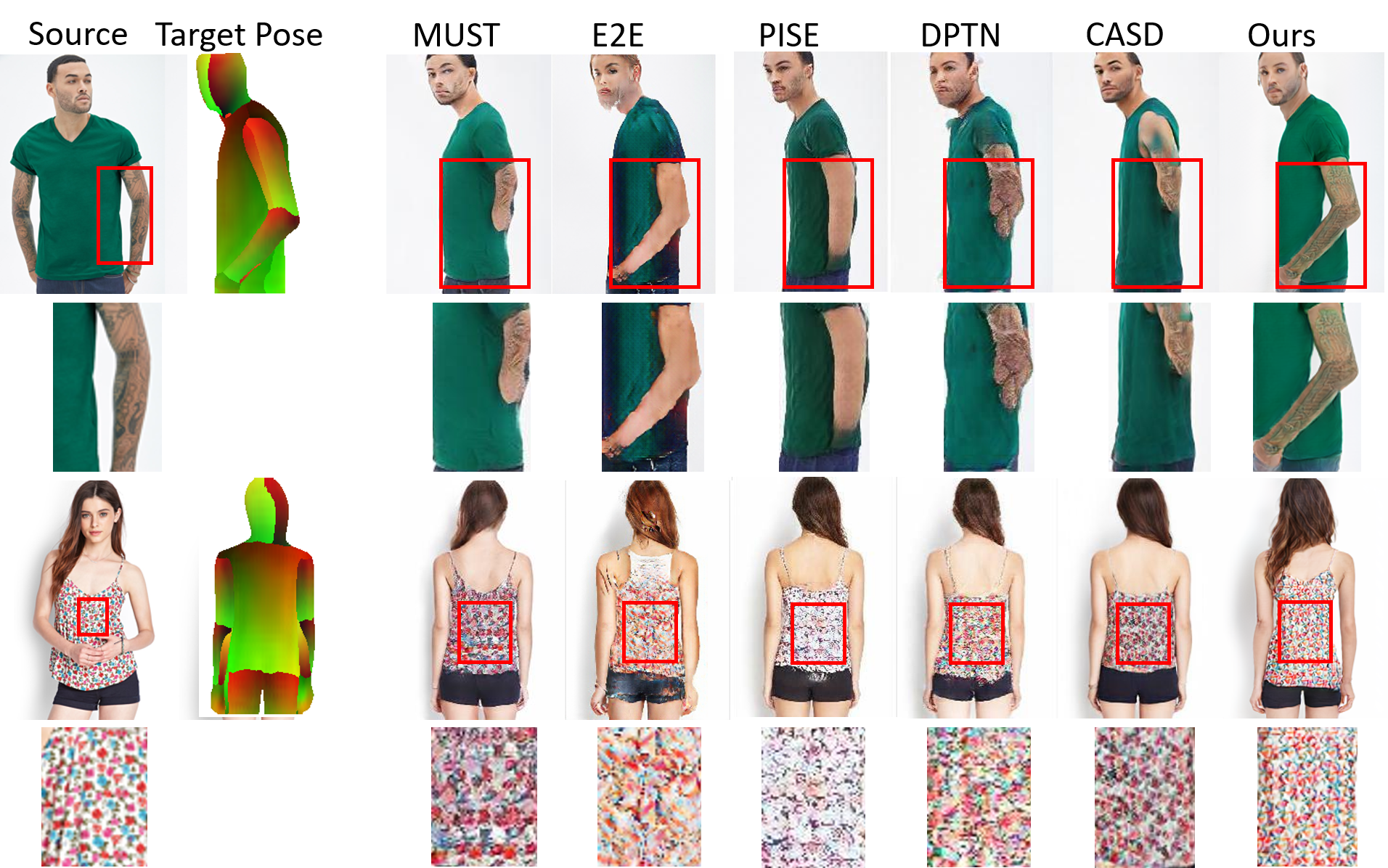}
    \label{fig:compare_deepf}
 \end{subfigure}
 \hfill
  \begin{subfigure}[c]{0.37\textwidth}
  \centering
    \includegraphics[width=\textwidth]{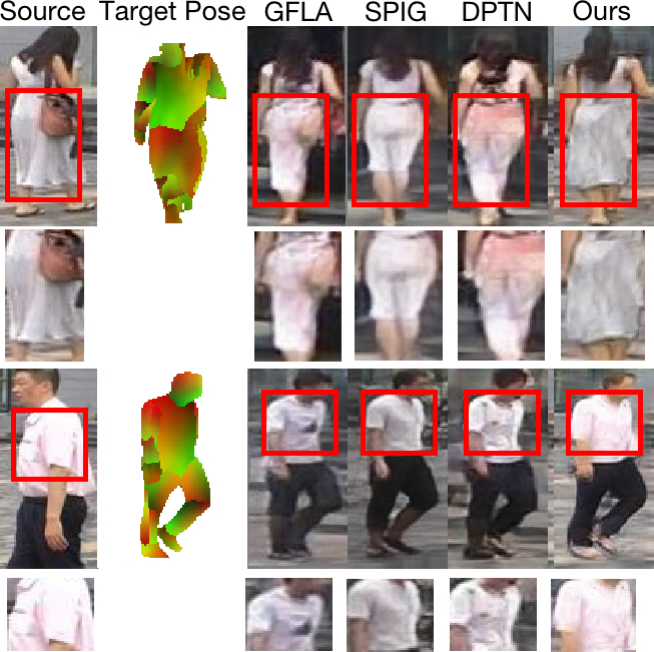}
    \label{fig:compare_market}
 \end{subfigure}
    \caption{Qualitative pose transfer results on DeepFashion (left) and Market-1501 (right). We enlarged the marked area for a better view of clothing details. MUST, E2E, and our \modelname{} are trained with unpaired data, while the rest are supervised by paired data. Our approach transfers the high-frequency texture patterns better than prior work.} 
 \label{fig:compare}
 \vspace{-2mm}
\end{figure*}

\begin{table*}[!t]
    \centering
    \begin{tabular}{lccccc}
    \toprule
     &\multicolumn{3}{c}{\textbf{Supervised by paired images}}
     &\multicolumn{2}{c}{\textbf{No paired images}}\cr  \cmidrule(r){2-4} \cmidrule(l){5-6}
        & PISE \cite{zhang2021pise} & DPTN \cite{zhang2022exploring}  &CASD \cite{zhou2022cross} &E2E \cite{song2019unsupervised}  &MUST \cite{ma2021must}  \cr 
        \midrule      
        Facial Identity Preservation   &53.2\%\small{$\pm$3.53} &65.1\%\small{$\pm$3.37} &50.4\%\small{$\pm$3.53} &76.9\%\small{$\pm$2.98} &69.2\%\small{$\pm$3.26}\cr
        Image Plausibility  &58.3\%\small{$\pm$3.49} &66.7\%\small{$\pm$3.33} &51.2\%\small{$\pm$3.53}
        &76.6\%\small{$\pm$2.99} &63.5\%\small{$\pm$3.40}\cr
        Texture Correctness  & 68.7\%\small{$\pm$3.27} & 66.2\%\small{$\pm$3.35} & 53.8\%\small{$\pm$3.53} &79.7\%\small{$\pm$2.84} &72.6\%\small{$\pm$3.15}  \cr 
        Average  &60.1\%\small{$\pm$3.46} &66.0\%\small{$\pm$3.27} &51.8\%\small{$\pm$3.53} &77.7\%\small{$\pm$2.93} &68.4\%\small{$\pm$3.29}\cr
         \bottomrule
    \end{tabular}   
        \caption{A/B user preferences on DeepFashion. We report how often our approach was selected as most like the reposed image with respect to facial identity, image plausibility, and texture correctness. The number that follows $\pm$ is the corresponding standard deviation. Our results are even preferred over some fully supervised methods} 
    \label{tab:user}   
\end{table*}

\noindent\textbf{Benefits of Using Unpaired Data.} In addition to requiring less annotation, our self-driven model can also be trained and directly fit to in-the-wild unpaired datasets while supervised methods can not. To simulate this scenario, we use NTED(Market) \cite{ren2022neural} trained on paired Market-1501 as our supervised model and test on DeepFashion as our in-the-wild dataset in Table \ref{tab:deepfashion}. Our PT$^2$ model was trained only on DeepFashion using unpaired data. Similarly, in Table \ref{tab:market}, NTED(DF) is trained on paired DeepFashion and tested on Market-1501 as the in-the-wild dataset. In Tables \ref{tab:deepfashion} \& \ref{tab:market}, PT$^2$ obtains better evaluation scores than NTED(Market) and NTED(DF) by a significant margin, which demonstrates an advantageous use case where unpaired training enables direct fitting on the target domain.
\smallskip

\noindent\textbf{User Study.} To verify the quality of generated images, we also conducted human evaluations on DeepFashion using Amazon Mechanical Turk. We collected 3 judgments for 50 images (150 total). Each worker was presented 3 pictures: the true image, a \modelname{} generated image, and an image generated by a method from prior work \cite{ma2021must,song2019unsupervised,zhang2021pise,zhang2022exploring,zhou2022cross}. The worker was asked to evaluate the quality of the generated images in three aspects: clothing texture correctness, facial identity preservation, and image plausibility. Table \ref{tab:user} shows that among self-driven methods, more than 68\% of workers believe our method achieves higher fidelity in the generated images in terms of clothing texture correctness, facial identity preservation, and image plausibility. Compared with approaches supervised by paired images, our method achieves comparable performance with an average of over 64\% user preference, demonstrating the effectiveness of our proposed approach.

\begin{figure}[!t]
\centering
\includegraphics[width=0.5\textwidth]{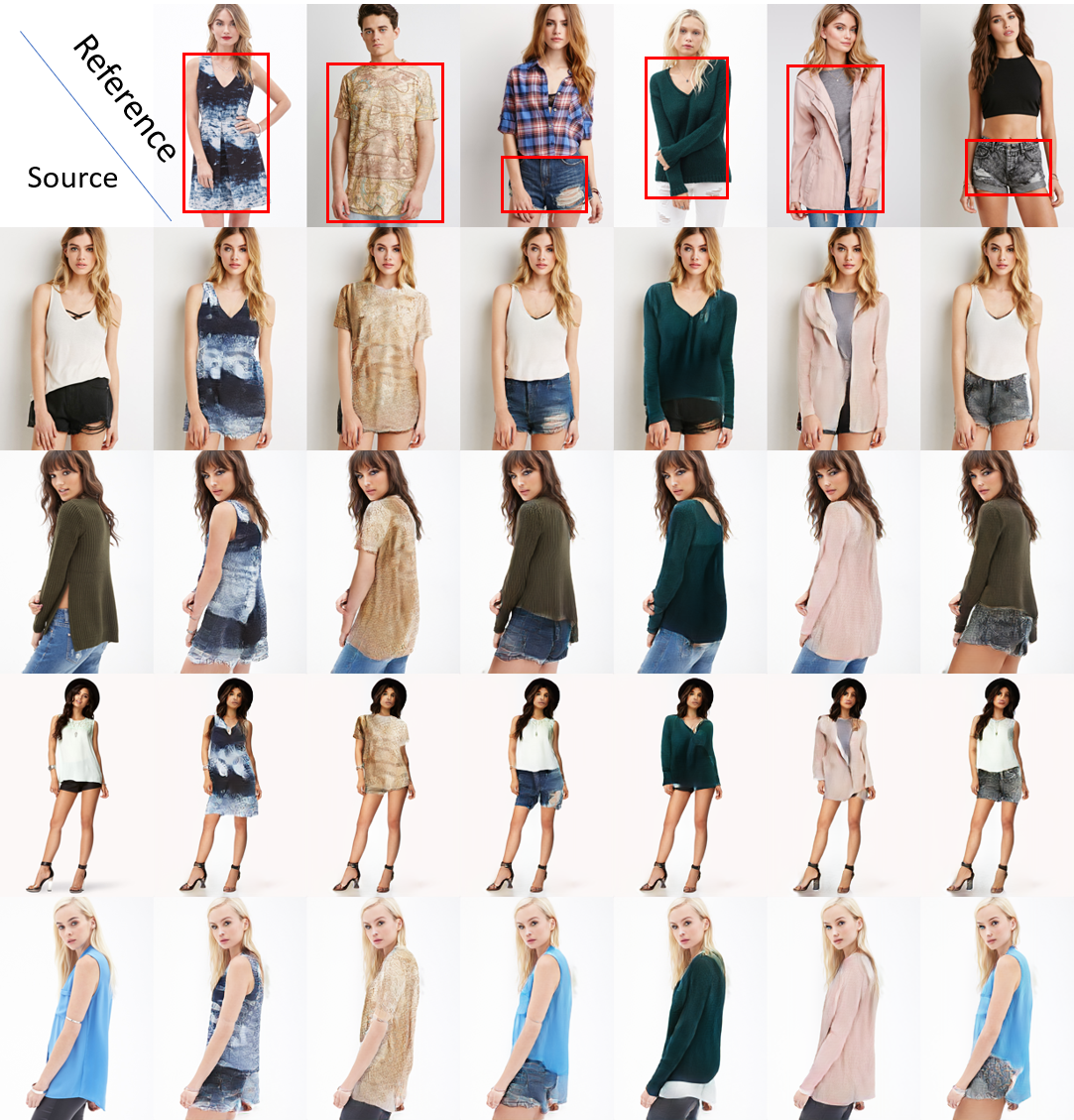}
\caption{Qualitative results of garment replacement. The left column is the source person and the top row is the reference(target) garment. The target garments are marked with red bounding boxes. Our approach captures the interactions within outfit configurations when transferring appearance.} 
\label{fig:gm_rp}
\vspace{-3mm}
\end{figure}
\subsection{Qualitative Results}
\noindent\textbf{Pose Transfer.} 
Fig.\ \ref{fig:compare} visualizes pose transfer results. We enlarged the area marked with a red bounding box for a better view of clothing details. In the first row (left), our method transferred the arm tattoos to the target pose while other methods either ignored this detail or failed to reconstruct the arm. Our model also reconstructs the color pattern in the second row (left) better than other approaches. This is likely due to the small-scale encoder in our model capturing detailed texture and thus reconstructing it based on the target pose. On the right side of Fig.\ \ref{fig:compare}, compared with supervised pose transfer methods, our approach faithfully recovered the shape and color of the dress and shirt in the two examples. More examples and failure analysis are included in the supplementary. 
\smallskip

\begin{table*}[!t]
    \centering
    \begin{tabular}{rlcccc}
    \toprule
    &Method &FID$\downarrow$ &SSIM$\uparrow$ &LPIPS$\downarrow$ &IS$\uparrow$  \cr \midrule  
    \textbf{(a)} &
    Input Warping, \textit{w/o.} Input Permuting &10.011 &0.781$\pm$0.072 &0.178$\pm$0.060 &\textbf{3.579$\pm$0.086}\cr
    &\textit{w/o.} Input Permuting (part-wise) &10.279 &0.780$\pm$0.069 &0.169$\pm$0.059 &3.525$\pm$0.095\cr  
    &\textit{w/o.} Pose Masking &9.451 &0.776$\pm$0.069 &0.176$\pm$0.059 &3.516$\pm$0.073\cr
    &\textit{w/o.} Texture Masking &8.780 &0.791$\pm$0.067 &0.164$\pm$0.059 &3.432$\pm$0.083\cr
    & \textit{w/o.} Parsing &8.460 &0.791$\pm$0.068 &0.161$\pm$0.159 &3.302$\pm$0.075 \cr
    \midrule     
    \textbf{(b)} &
     \textit{w/o.} small kernel &8.905 &0.782$\pm$0.067 &0.170$\pm$0.060 &3.442$\pm$0.119\cr
     &\textit{w/o.} large kernel  &9.275 &0.785$\pm$0.068 &0.166$\pm$0.060 &3.401$\pm$0.078\cr
     & Patch Concat &8.514 &0.784$\pm$0.068 &0.168$\pm$0.059 &3.459$\pm$0.090\cr
     \midrule 
      \textbf{(c)} 
    &Pose Concat &9.236 &0.775$\pm$0.067 &0.170$\pm$0.059 &3.438$\pm$0.081\cr
    & \textit{w/o.} Texture Matching &10.786 &0.787$\pm$0.067 &0.172$\pm$0.060 &3.489$\pm$0.085 \cr
      \midrule
      &\modelname{} + Input Warping & \textbf{8.332} &0.792$\pm$0.059 &0.160$\pm$0.060 &3.470$\pm$0.090 \cr
     &\textbf{\modelname(Ours)} &8.338 &\textbf{0.795$\pm$0.067} &\textbf{0.158$\pm$0.059} &3.469$\pm$0.098 \cr
    \bottomrule
    \end{tabular}
    \caption{Ablation study on DeepFashion reporting the contribution of each component of our model \modelname{}.}
    \label{tab:abl}
\end{table*}

\noindent\textbf{Garment Replacement.} Our approach can also be used to swap clothing pieces between people. Due to differences in the shape of the source and target garments (\eg, jeans and shorts), the model can synthesize the texture from the source garment in various ways on the target.
For instance, in Fig. \ref{fig:gm_rp}, we can see that the person in the third row has an untucked green shirt, which is tucked in when applying the shape of the shorts in the last column. 
Similarly, the shorts in the second row are occluded by the camel t-shirt and pink jackets, but visible in other columns. Overall, this demonstrates that the model captures the interactions within outfit configurations, using this to synthesize plausible outputs.



\subsection{Ablation Study}
\label{sec:abl}
We evaluate the effectiveness of each component of our model in an ablation study. More part-wise analyses are provided in Sec.\ A of the supplementary.
\smallskip

\noindent\textbf{Effects of input permutation.} In Table \ref{tab:abl}(a), \textit{\textit{w/o.} Input Permuting (part-wise)} uses a part-wise encoder but does not permute the inputs, which results in entangled pose and textures. \textit{Input Warping, w/o. Input Permuting} uses Thin Plate Spline (TPS) transformation to warp the source image, which can be viewed as a mild way of disentangling the pose and texture at the image-level \cite{lorenz2019unsupervised}. \textit{\textit{w/o.} Pose Maksing} does not mask the source pose, which could cause the model to overfit to visible pose regions. \textit{\textit{w/o.} Texture Masking} does not mask the textures in the texture branch. \textit{\textit{w/o.} Parsing} removes the parsing map in the input. As shown in Table \ref{tab:abl}, our complete model \modelname{} improves all the metrics, demonstrating the effectiveness of the proposed input permuting function. To further verify that the arm/body orientation in the small patches will not fail the disentanglement, we combine TPS warping with input permutation using large rotation and scaling in \modelname{} + Input Warping.  Our results show no additional improvement, suggesting significant disentanglement is already being achieved by our permutation function. 
\smallskip

\noindent\textbf{Effects of using dual-scale encoder.}  In Table \ref{tab:abl}(b), \textit{\textit{w/o.} small kernel} uses only the large kernel in the feature encoders, which leads to a loss of clothing detail. \textit{\textit{w/o.} large kernel} uses only the small kernel encoder, degrading the model's capacity for long-range texture patterns. The baseline approach \textit{Patch Concat} avoids the boundary issue by encoding each patch separately, limiting the receptive field to be at most the patch size. This approach performs well, but the dual-scale approach still yields the best overall performance. 
\smallskip

\noindent\textbf{Effects of using three branches.} In Table \ref{tab:abl}(c), \textit{Pose Concat} removes the source pose branch and concatenates the source pose representation channel-wise to the texture input. Although \textit{Pose Concat} has provided the geometry information for the permuted source pose, not directly learning the geometric transformation between postures through separate branches still leads to a significant drop in perceptual metrics. We also provide the results of \textit{w/o.\ Texture Match}, which removes the texture-to-pose match in our cross-attention module (Fig.\ \ref{fig:att}). Without texture match, pose match alone is not sufficient to learn the spatial support of clothing. Consider the task of transferring a flared skirt: much of the skirt's surface cannot be directly associated with a position on the human pose map. Therefore, adding the texture-to-pose correspondence (texture match) can help the model learn this type of spatial support. As reported in Table \ref{tab:abl}, our full model \modelname{} that incorporates the texture match shows better performance on all perpetual metrics.
\smallskip

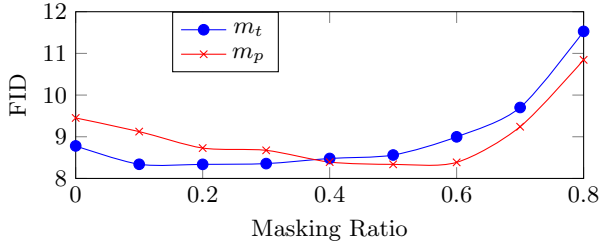
\begin{figure}
    \centering
    \small
\begin{tikzpicture}
    \begin{axis}[
    width=\columnwidth,
    height=3.8cm,
    xlabel near ticks,
    ylabel near ticks,
        ylabel=FID,
        xlabel=Masking Ratio,
        xmin=0, xmax=0.8,
        ymin=8, ymax=12,
        legend style={at={(0.4,1),anchor=south west}}
        ]
    \addplot[smooth,mark=*,blue] plot coordinates {
    (0.0, 8.78)
    (0.1, 8.340)
    (0.2, 8.338)
    (0.3, 8.356)
    (0.4, 8.479)
    (0.5, 8.562)
    (0.6, 8.998)
    (0.7, 9.702)
    (0.8, 11.528)
    };
    \addlegendentry{$m_t$}
    \addplot[smooth,color=red,mark=x]
        plot coordinates {
        (0.0, 9.451)
        (0.1 ,  9.124)
        (0.2  , 8.729)
        (0.3   ,8.674)
        (0.4 ,  8.393)
        (0.5 ,  8.338)
        (0.6 ,  8.386)
        (0.7 ,  9.241)
        (0.8 , 10.845)
           };
    \addlegendentry{$m_p$}
    \end{axis}    
    \end{tikzpicture}
    \vspace{-2mm}
    \caption{Masking ratio vs.\ FID. We set $m_p=0.5$ when varying $m_t$, and $m_t=0.2$ when varying $m_p$. The masked patches help the model learn how to reconstruct occluded regions.}
    \label{fig:mask}
\end{figure}

\begin{table}[t]
    \centering
\setlength{\tabcolsep}{1.5pt}    
   \begin{tabular}{lcccc}
    \toprule      
    Patch &\multirow{2}{*}{FID$\downarrow$} &\multirow{2}{*}{SSIM$\uparrow$} &\multirow{2}{*}{LPIPS$\downarrow$} &\multirow{2}{*}{IS$\uparrow$}  \cr 
    Size \cr\midrule  
     8 &8.565 &0.789\small{$\pm$0.068} &0.161\small{$\pm$0.059} &3.386\small{$\pm$0.101}\cr
     16 &\textbf{8.338} &\textbf{0.795\small{$\pm$0.067}} &\textbf{0.158\small{$\pm$0.059}} &3.469$\pm$0.098 \cr
     32 &8.401 &0.784\small{$\pm$0.067 } &0.168\small{$\pm$0.058} &3.503\small{$\pm$0.100}\cr
     64 &8.898 &0.776\small{$\pm$0.640} &0.169\small{$\pm$0.056}
     &\textbf{3.555\small{$\pm$0.139}}\cr  
      \bottomrule
    \end{tabular}
    
    \caption{Ablations over patch sizes on DeepFashion. We find patch size 16 works best for our model.}
    \label{tab:psize}
    \vspace{-2mm}
\end{table}

\noindent\textbf{Effects of patch size.} The amount of pose information within a patch is directly related to its size, which can affect our model's performance. Table \ref{tab:psize} reports the effect of varying the patch size of our image permutation function from Sec.\ \ref{sec:ip}. We find that smaller patches improve disentanglement, but result in more boundary-crossings. Larger patch sizes have fewer crossings, but results in more entangling of pose and texture. Table \ref{tab:psize} shows that a patch size of 16 reflects the best balance of these two factors.
In addition, we observe that even the suboptimal large patch sizes does not deteriorate downstream task performance too significantly. This arises due to the high masking rate of 16$\times$16 patches in both the image and source pose in addition to the permutations (Sec.\ \ref{sec:ip}). Thus, the network is forced to learn to sample correct textures for the masked pixels based on the target pose, helping to avoid an identity mapping of the input person when the patch size is closer to the image size.
\smallskip

\noindent\textbf{Ablations on masking ratios.} Patch masking helps the model learn to reconstruct occluded regions. Fig.\ \ref{fig:mask} illustrates this effect by varying the masking ratios of the source pose branch ($m_p$) and the texture branch ($m_t$), respectively. We find that when $0.1\le{m_t}\le0.3$ and $0.4\le{m_p}\le0.6$, the results are mostly flat, demonstrating that we can obtain gains with minimal tuning. However, lower values cause the model to overfit to visible patches, and higher values hurts performance due to many missing texture/pose details.

\section{Conclusion}
We propose \modelname, a self-driven human pose transfer method that randomly permutes image patches, helping to disentangle pose from texture, then reconstructs the original image from the permuted input. The approach helps to achieve patch-level disentanglement and detail-preserving texture transfer. Thus, \modelname is capable of more accurately reconstructing garment textures in pose transfer tasks than prior work. 
We also show that using a dual-scale encoder can be resilient to permutation noise while aligning pose and textures. Experiments on DeepFashion and Market-1501 show that our model improves the image quality of self-driven approaches, with higher user preference scores than prior work.
Moreover, \modelname obtains comparable objective and subjective results to most pose transfer methods supervised by paired data.
\smallskip

\noindent\textbf{Acknowledgements} This material is based upon work supported, in part, by DARPA under agreement number HR00112020054 and the National Science Foundation under Grant No.\ DBI-2134696. Any opinions, findings, and conclusions or recommendations expressed in this material are those of the author(s) and do not necessarily reflect the views of the supporting agencies.


{\small
\bibliographystyle{ieee_fullname}
\bibliography{refs}
}

\renewcommand{\thesection}{\Alph{section}}
\setcounter{section}{0}

\twocolumn[{%
\renewcommand\twocolumn[1][]{#1}%
\centering{\huge Supplementary}
\vspace{1cm}
\begin{center}
\includegraphics[width=\textwidth]{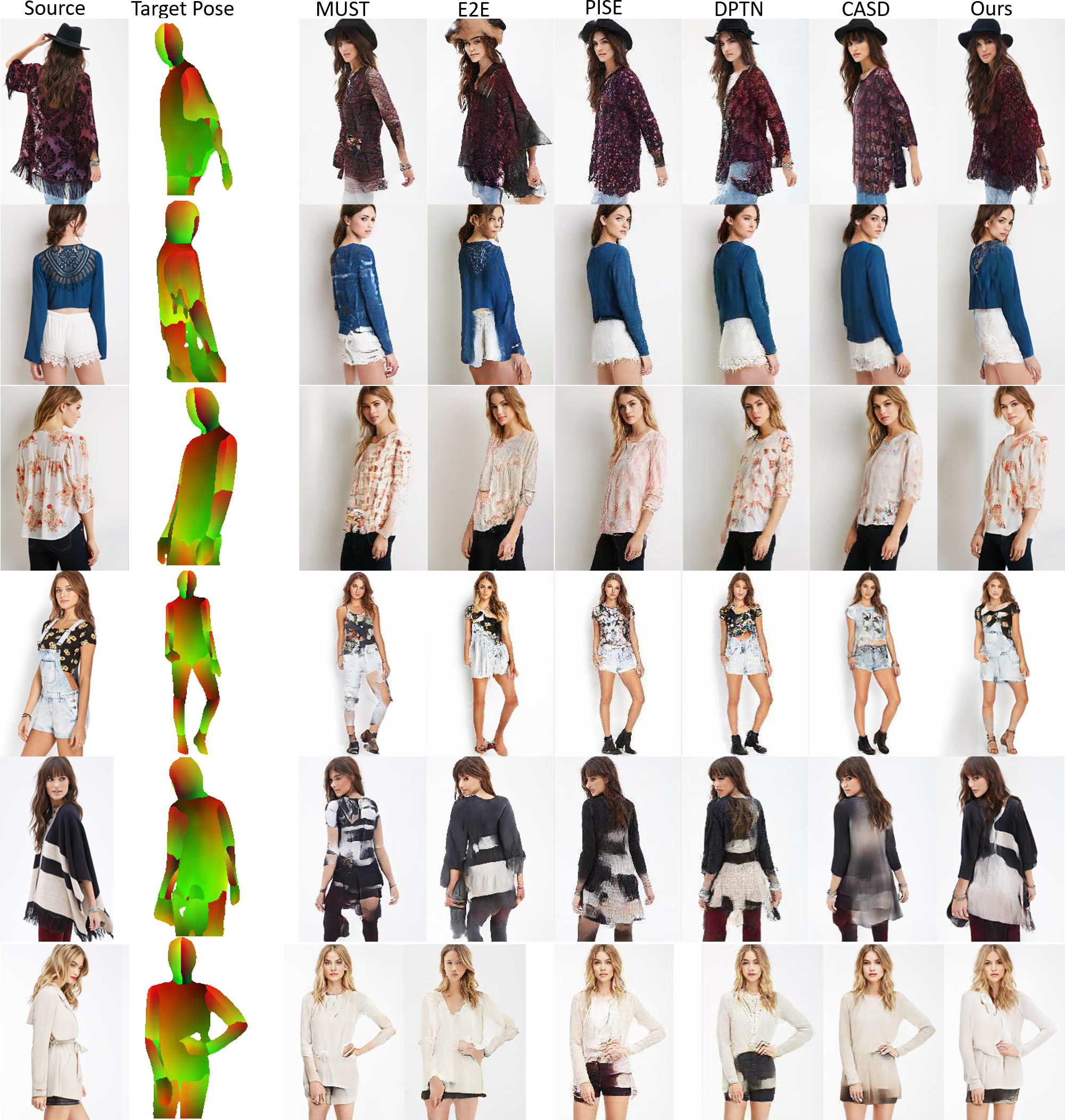}
    \captionof{figure}{Additional pose transfer examples on DeepFashion. MUST, E2E, and ours are trained with unpaired images. Other methods are supervised by paired data. }  
    \label{fig:more_deepf}
\end{center}
}]
\twocolumn[{%
\renewcommand\twocolumn[1][]{#1}%
\maketitle
\begin{center}
\includegraphics[width=\textwidth]{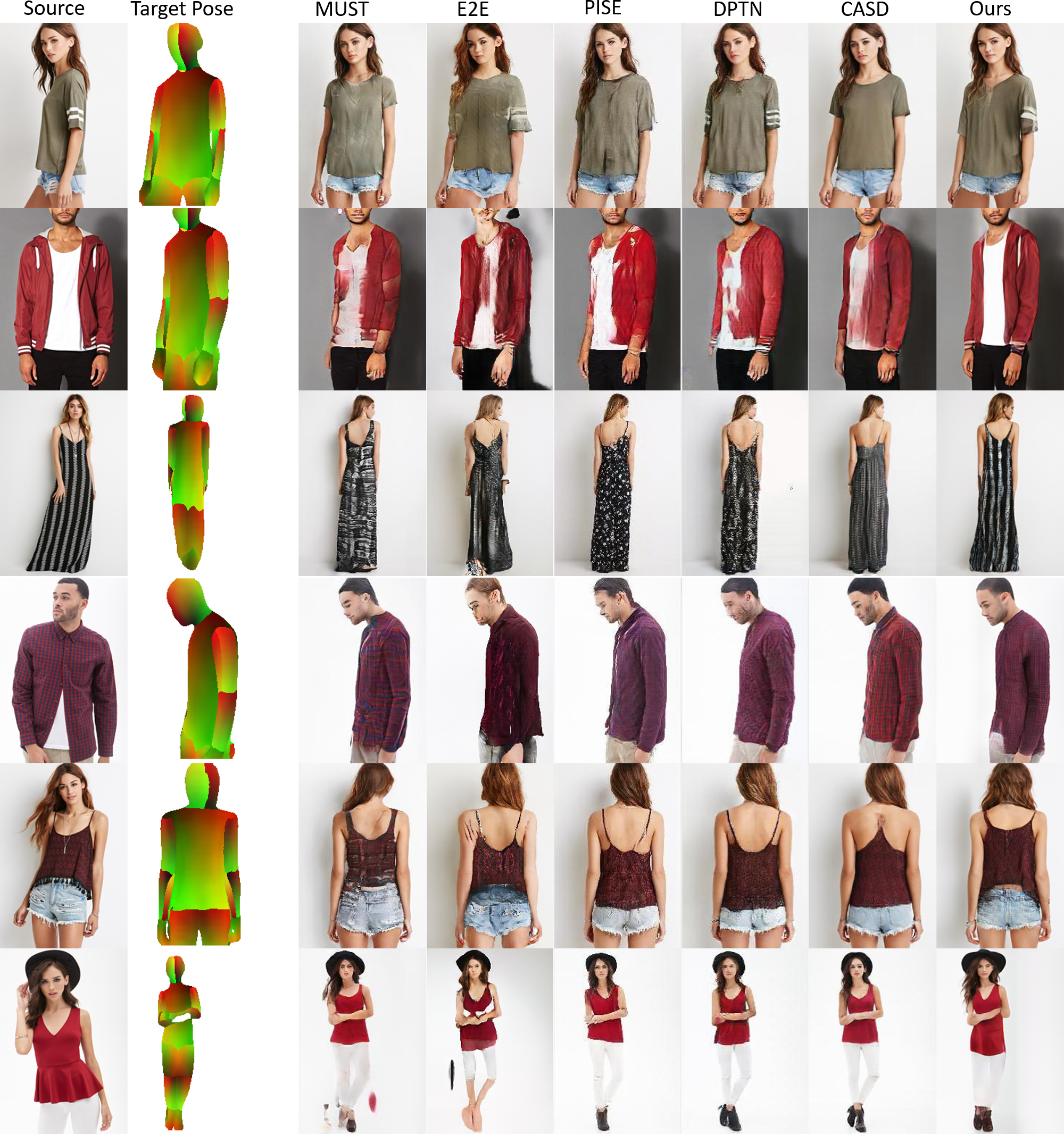}
    \captionof{figure}{Additional pose transfer examples on DeepFashion. MUST, E2E, and ours are trained with unpaired images. Other methods are supervised by paired data. }  
    \label{fig:more_deepf2}
\end{center}
}]
\twocolumn[{%
\renewcommand\twocolumn[1][]{#1}%
\maketitle
\begin{center}
\includegraphics[width=\textwidth]{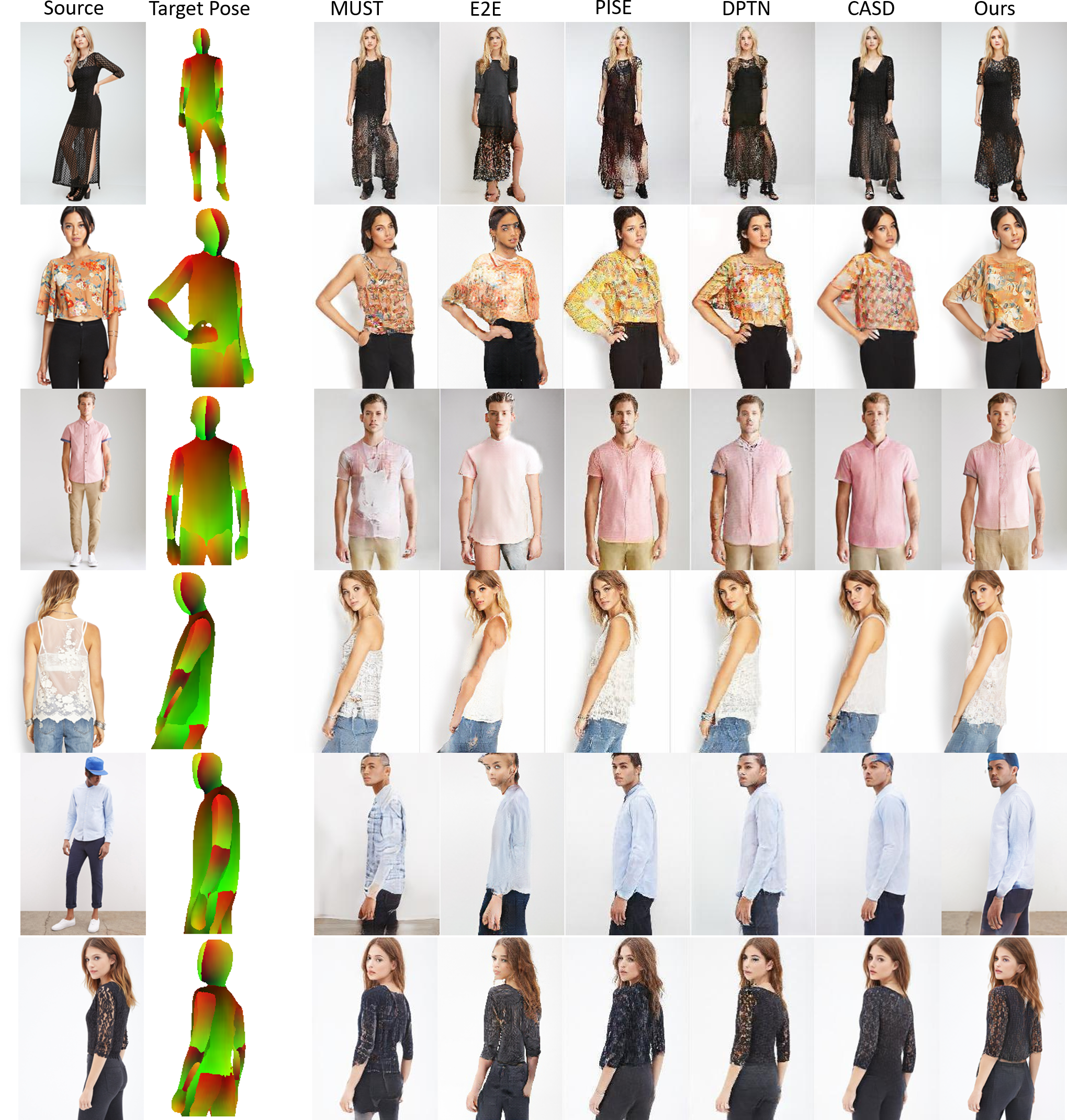}
    \captionof{figure}{Additional pose transfer examples on DeepFashion. MUST, E2E, and ours are trained with unpaired images. Other methods are supervised by paired data. }  
    \label{fig:more_deepf3}
\end{center}
}]

\twocolumn[{%
\renewcommand\twocolumn[1][]{#1}%
\vspace{-1cm}
\begin{center}
\includegraphics[width=\textwidth]{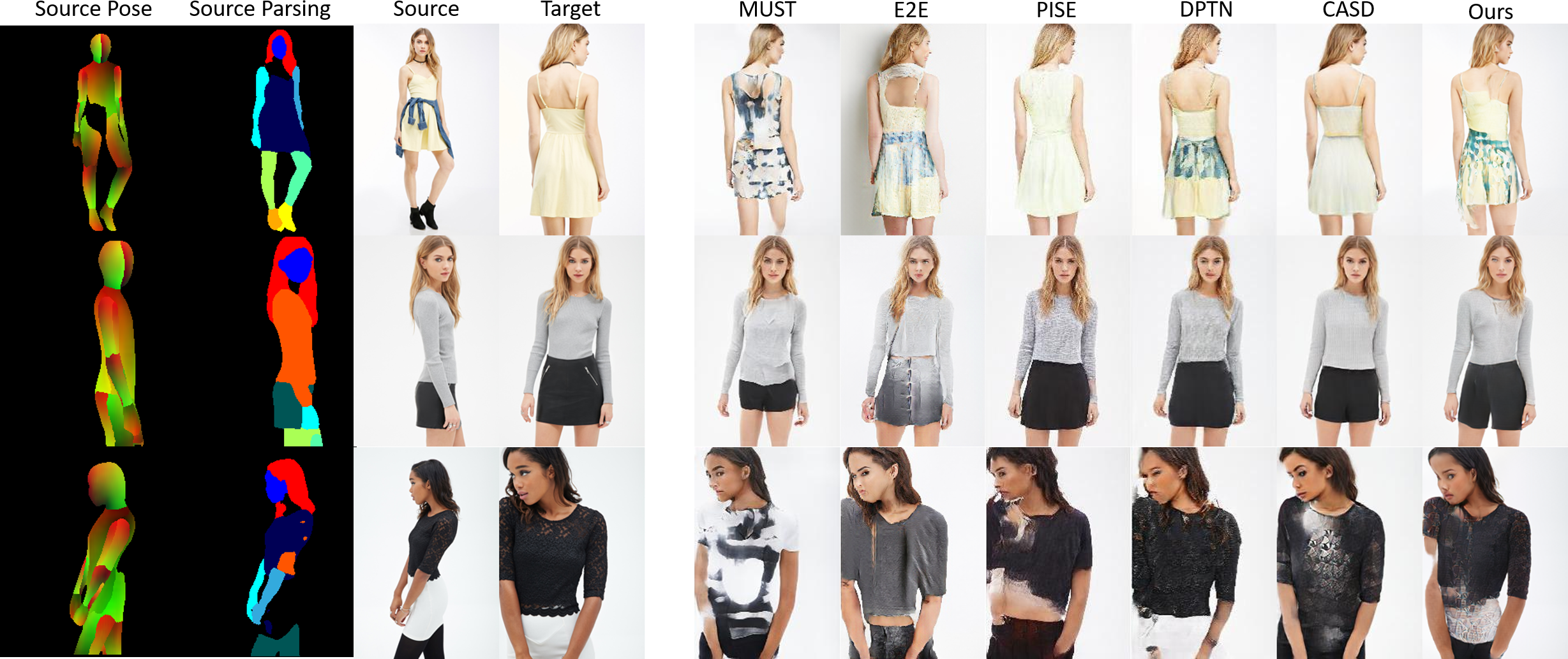}
\vspace{-0.5cm}
    \captionof{figure}{Failure cases in DeepFashion. Many failures are due to incorrect predictions of the source UV map and source parsing map.}  
    \label{fig:fail}
\end{center}
\begin{center}
\small
    \begin{tabular}{llcccccccc}
    \toprule
    &&\multicolumn{4}{c}{SSIM} &\multicolumn{4}{c}{IoU} \cr \cmidrule(lr){3-6} \cmidrule(lr){7-10}
    &&Head &Clothes &Arms &Legs &Head &Clothes &Arms &Legs \cr \midrule
     \textbf{(a)} &\textit{w/o.} Input Permuting (part-wise) &0.299 &0.349 &0.377 &0.407 &0.682 &0.772 &0.642 &0.509\cr
     &Input Warping, \textit{w/o.} Input Permuting  &0.298 &0.351 &0.376 &0.408 &0.697 &0.782 &0.641 &0.525\cr
      &\textit{w/o.} Pose Masking  &0.298 &0.343 &0.343 &0.382 
      &0.674 &0.765 &0.625 &0.502\cr
     &\textit{w/o.} Texture Masking &0.351 &0.369 &0.421 &0.414
     &0.697 &0.777 &0.667 &0.507\cr
     &\textit{w/o.} Parsing &0.353 &0.371 &0.423 &\textbf{0.441}
     &0.703 &0.787 &0.673 &\textbf{0.552} \cr
    \midrule
    \textbf{(b)} &\textit{w/o.}small kernel &0.305 &0.368 &0.394 &0.395
    &0.674 &0.778 &0.649 &0.505\cr
    &\textit{w/o.} large kernel &0.335 &0.365 &0.396 &0.407
    &0.689 &0.783 &0.653 &0.522 \cr
    &Patch Concat &0.331 &0.365 &0.398 &0.407
    &0.686 &0.778 &0.653 &0.525\cr
    &\textit{w.} blur &0.239 &0.354 &0.310 &0.343 &0.625 &0.784 &0.601 &0.454\cr
    \midrule
    \textbf{(c)} &\textit{w/o.} Source Pose Branch &0.348 &0.371 &0.433 &0.361
    &0.706 &0.780 &0.647 &0.428\cr
    &Pose Concat &0.321 &0.364 &0.425 &0.371 &0.709 &0.784 &0.648 &0.433\cr
    \midrule
    \textbf{(d)} &E2E \cite{song2019unsupervised} &0.362 &0.307 &0.337 &0.350 &0.709 &0.761 &0.635 &0.501\cr
    &MUST \cite{ma2021must} &\textbf{0.371} &0.312 &0.404 &0.297 &0.707 &0.737 &0.559 &0.424 \cr    
    \midrule
    &Ours &\textbf{0.371} &\textbf{0.377} &\textbf{0.452} &0.428
    &\textbf{0.717} &\textbf{0.791}  &\textbf{0.688} &0.531\cr
    \bottomrule
    \end{tabular}
    \captionof{table}{Part-wise scores in DeepFashion. (a), (b) and (c) are our ablations. (d) includes self-driven methods trained without paired data.}
    \label{tab:ssim_part}
\end{center}
}]


\begin{figure*}[!t]
  \centering
    \begin{subfigure}{\textwidth}
    \includegraphics[width=\textwidth]{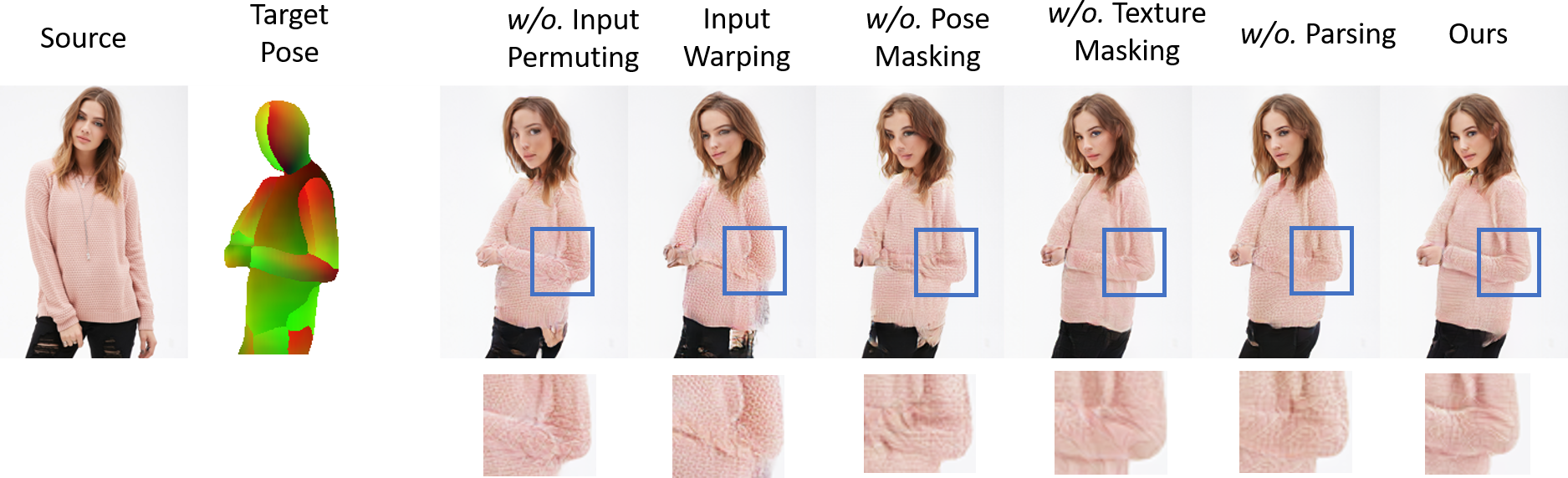} 
    \caption{Examples of ablations on the input permutation function.}
    \label{fig:abl1}
    \end{subfigure}
    \begin{subfigure}{0.7\textwidth}
    \includegraphics[width=\textwidth]{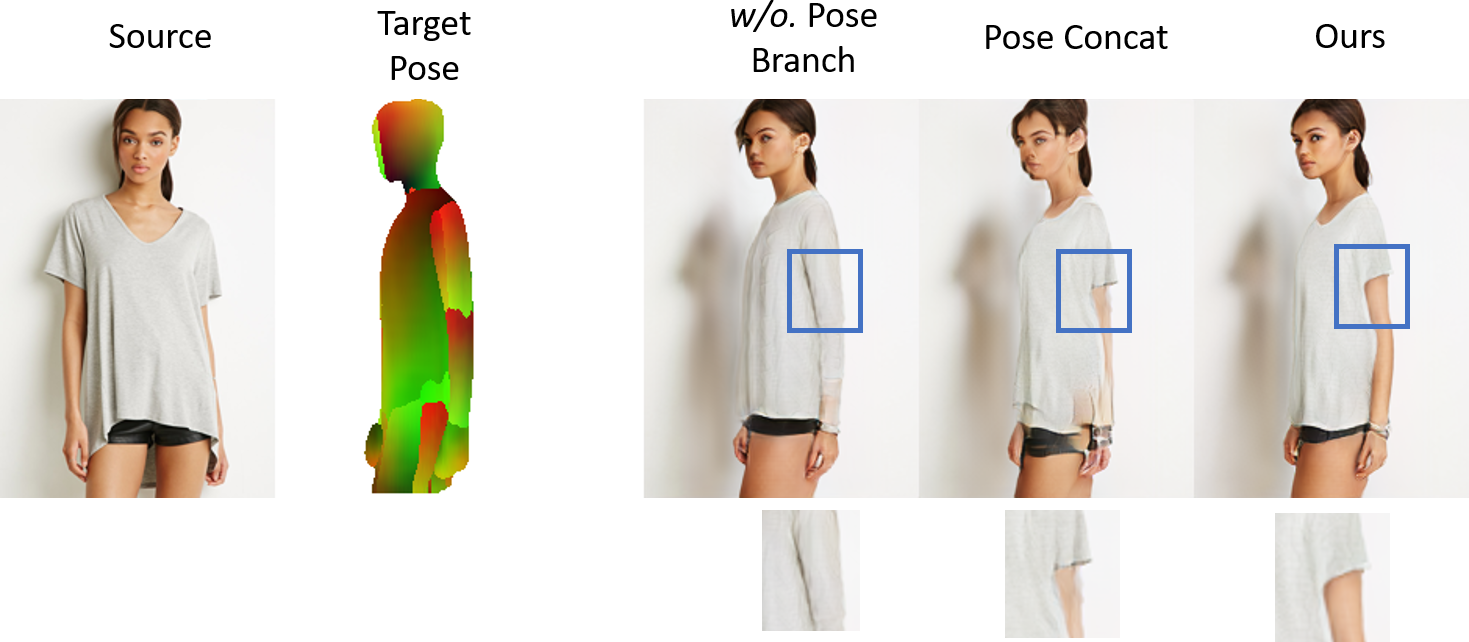} 
    \caption{Examples of ablations on the source pose branch.}
     \label{fig:abl2}
    \end{subfigure}
    \begin{subfigure}{0.9\textwidth}
    \includegraphics[width=\textwidth]{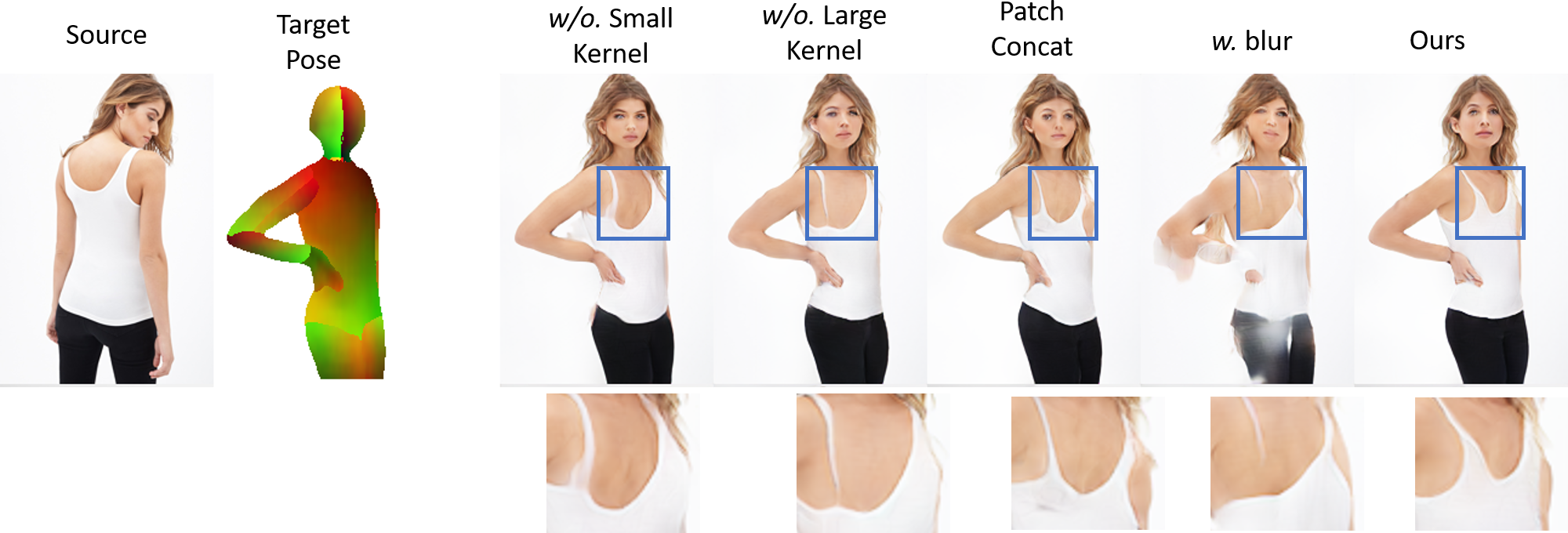} 
    \caption{Examples of ablations on the dual kernel encoder.}
     \label{fig:abl3}
    \end{subfigure}
    \caption{Generated images of ablations of our model. Each component of our model improves the transfer of shape information and detailed clothing patterns, resulting in our full model obtaining the best results.} 
    \label{fig:abl}
 \end{figure*}

\section{Discussions}
\label{sec:discuss}
\noindent\textbf{Failure case analysis.} Figures \ref{fig:more_deepf}-\ref{fig:more_deepf3} provide several successful examples generated by the proposed method on DeepFashion. We also obtain the part segments of the images in the test set using the human parser in \cite{Zhang_2020_CVPR}, and then compute their part-wise scores. Table \ref{tab:ssim_part}(d) shows that our model achieves better texture transfer and shape reconstruction than prior work \cite{song2019unsupervised,ma2021must} in terms of part-wise SSIM and IoU. However, one limitation of our model is that it relies on the segmentation map and DensePose prediction of the source image to obtain semantic and position information for the permuted textures. We found the accuracy of the offline human parser and DensePose model greatly affects the transfer results. Figure \ref{fig:fail} shows several failed examples due to this type of inaccuracy. In the first row, the coat wrapped around the dress was misclassified as part of the dress in the parsing map, for which our generated back view incorrectly mixes up their textures. Similarly, the skirt in the second row was classified as shorts in the parsing map. As a result, our generator infers the occluded clothing piece as shorts in the front view. In the last row, the color of the skirt is half-black and half-white because the skirt piece was not identified in the parsing map.

 \begin{figure*}[!t]
     \centering
     \includegraphics[width=0.9\textwidth]{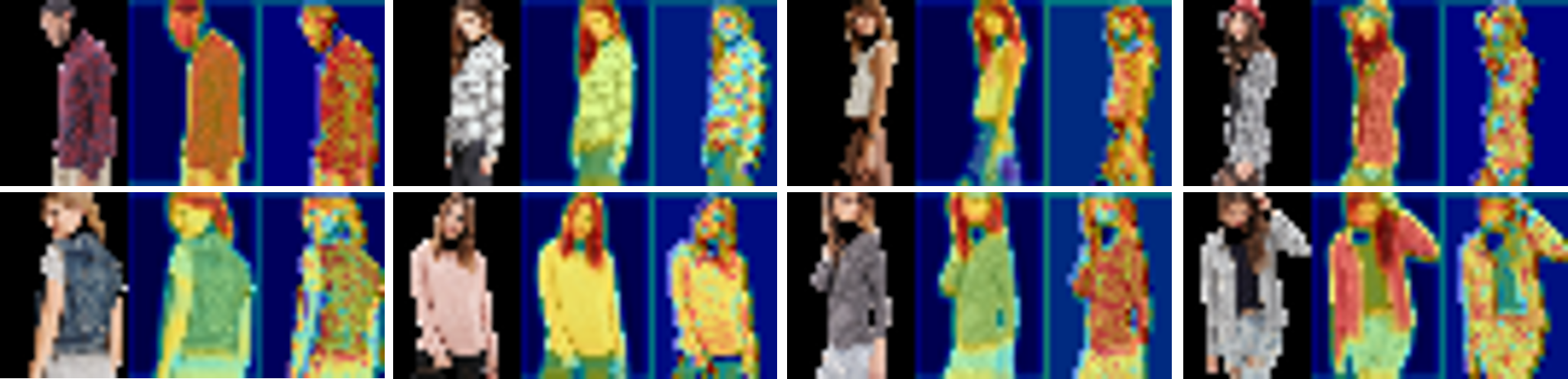}
     \caption{Visualized feature map of the encoded texture features. The feature map is overlaid with the source image. The source image is downsampled to the resolution of the feature map. Each triplet includes a downsampled source image, the feature map from the large-kernel encoder, and the feature map from the small-kernel encoder. Red indicates a higher value and blue means a smaller value.}
     \label{fig:feature}
 \end{figure*}

\noindent\textbf{Analysis on the ablations.} We present the part-wise scores in Table \ref{tab:ssim_part} and some visualized examples in Figure \ref{fig:abl} to show the functionality of each component of our model. IoU in Table \ref{tab:ssim_part} means the Intersection over Union score between the segmentation maps of the generated image and that of the target image. This metric evaluates the shape consistency of each body part after pose transfer. 

From Table \ref{tab:ssim_part}(a), we find that \textit{w/o. Input Permuting} and \textit{Input Warping} have larger drops on arms and heads compared to other body parts, which indicates their incapability of transferring human posture/shape. For example, in Figure \ref{fig:abl1}, both \textit{w/o. Input Permuting} and \textit{Input Warping} result in a distorted face and obvious edge blurring at the elbow after pose transfer. The ablation model \textit{w/o. Pose Masking} overfits to an identity mapping function between the source and target pose, leading to the lowest IoU on all body parts in Table \ref{tab:ssim_part}(a). The ablation model \textit{w/o. Texture Masking} is much better than \textit{w/o. Pose Masking} since not masking the texture can still get a correct pose transformation function from the source pose branch. \textit{w/o. Parsing} has lower scores on most body parts compared to the full model, suggesting that including a parsing map in the input is overall beneficial to the pose transfer task. 

In Table \ref{tab:ssim_part}(b), \textit{w/o. small kernel} has worse performance than \textit{w/o. large kernel} and \textit{Patch Concat}, indicating the importance of the small kernel encoder in reducing noise caused by input permutation. \textit{w/o. large kernel} and \textit{Patch Concat} show similar part-wise scores since their receptive fields are respectively limited by the small kernel size and patch size. As shown in Figure \ref{fig:abl3}, without the small-kernel encoder, the ablation model correctly transfers color, but blurs the edges of the strap. In \textit{w/o.} large kernel and \textit{Patch Concat}, the model has a limited receptive field and thus fails to recover the exact shape of the strap. To see if the large-kernel encoder is learning certain low-level information (\eg color and shape) from permuted patches, we also tried replacing the inputs of the large-kernel encoder with heavily Gaussian blurred image without permutation (denoted by \textit{w.} blur). From both Table \ref{tab:ssim_part}(c) and the example in Figure \ref{fig:abl3}, we can see that images generated by \textit{w.} blur are much worse compared to those of the full model. This suggests that features learned by the large-kernel encoder from the permuted image might include high-frequency information that is lost in the Gaussian blurred texture.   

In Table \ref{tab:ssim_part}(c), concatenating the source pose representation with the texture slightly improves IoU, but does not improve SSIM. This means merging the source pose and texture in one branch can provide some shape information, but does not have the ability to match the precise relative position between the source and target pose. For example, in Figure \ref{fig:abl2}, although \textit{Pose Concat} reconstructs the short sleeve after pose transfer, it has artifacts around the edges of the arm. This is why our full model uses separate source pose and texture branches. By separating the source pose and texture, the source pose branch can directly learn the texture-agnostic geometry transformation between the two poses, and thus better recovers the shape.

To further explore the differences in texture features learned by the large-kernel encoder and the small-kernel encoder, we sum up the encoded feature maps across all channels in the texture branch, and normalize their values to be in the range $[0,1]$. Next, we downsample the image to the resolution of the feature map and overlay the normalized feature map with the downsampled source image. In Figure \ref{fig:feature}, each triplet includes the downsampled source image, the feature map from the large-kernel encoder, and the feature map from the small-kernel encoder. The feature map given by the large-kernel encoder (middle image in each triplet) appears to be much smoother than that of the small-kernel encoder (right image in each triplet). This suggests that the large-kernel encoder might be learning coarse information from the clothing piece (\eg, color and shape), while the small-kernel encoder is learning more fine-grained patterns (\eg, stripe and pleat).

\section{Training Details.} 

\begin{table*}
\centering
\begin{tabular}{lcccccc} 
    \toprule
    Method &FID$\downarrow$ &SSIM$\uparrow$ &M-SSIM$\uparrow$  &LPIPS$\downarrow$ &M-LPIPS$\downarrow$ &IS$\uparrow$ \cr \midrule 
    \multicolumn{7}{l}{
    \textbf{Supervised by paired images}}
    \cr 
    GFLA \cite{ren2020deep} &20.194 &0.286 &0.815 &0.274 &0.138 &2.546 \cr  
    SPIG \cite{lv2021learning} &22.043 &\textbf{0.317} &0.819 &\textbf{0.271} &0.129 &2.761 \cr
    DPTN \cite{zhang2022exploring} &17.929 &0.289 &0.820 &0.266 &0.125 &2.479 \cr
    NTED(DF) \cite{ren2022neural} &38.831 &0.191 &0.734 &0.353 &0.212 &2.242 \cr
    \midrule
    \multicolumn{7}{l}{
    \textbf{No paired images}} \cr   
    \modelname(Ours) &\textbf{17.389} &0.280 &\textbf{0.820} &0.314 &\textbf{0.122} &\textbf{2.789}\cr
    \bottomrule
    \end{tabular}
    \caption{Pose transfer results for $128 \times 64$ resolution images on Market-1501.}
    \label{tab:market}
\end{table*}
\begin{figure}[t]
    \centering
    \includegraphics[width=0.3\textwidth]{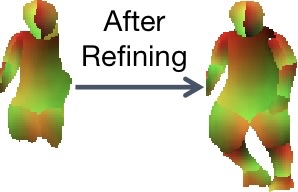}
    \caption{An example of the refined DensePose in Market-1501.}
    \label{fig:dpose}
\end{figure}

 \label{sec:training_details}
 We use AdamW optimizer \cite{loshchilov2018decoupled} for training with $\beta_1=0.5, \beta_2=0.999$. The initial learning rate is set to $10^{-3}$ and decays to $2\times 10^{-4}$ after five starting epochs. The trade-off parameters are set to $\lambda_1 =2.0, \lambda_2 =5.0, \lambda_3 =0.5, \lambda_4 =150$ in all experiments. The patch size is 16 $\times$ 16 for DeepFashion and 8 $\times$ 8 for Market-1501. To stabilize the training, we use the EMA strategy \cite{yaz2018unusual} to average the learned weights of the generator. We train on 256 $\times$ 176 images in DeepFashon and $128$ $\times$ 64 images on Market-1501. Our pose representation is predicted by DensePose \cite{guler2018densepose} and the parsing maps are obtained from CorrPM \cite{Zhang_2020_CVPR}. We found that the predicted dense pose in Market-1501 has poor quality as the image resolution is too low ($128\times 64$) for the DensePose model. Therefore, we use an offline super-resolution model \cite{liang2021swinir} to upsample the Market-1501 images to $512\times 256$, get dense pose from these images, and then downsample the pose to the original image resolution for our pose transfer task. An example of the refined DensePose representation is shown in Figure \ref{fig:dpose}. We also add human keypoints predicted from OpenPose \cite{openpose} as part of the pose representation to improve the accuracy of predicted posture on Market-1501.


\section{Analysis on Market-1501}

As shown in Table \ref{tab:market}, although our model outperforms fully-supervised approaches on M-SSIM and M-LPIPS, we note that we do perform worse according to SSIM and LPIPS, which is computed over the entire image rather than just the target person region.

 To investigate the reason behind the discrepancy when we use masked regions for evaluation, we computed part-wise SSIM scores. The scores for background, arms, legs, clothes, and head for PT$^2$ are: 0.237, 0.263, 0.283, 0.323, 0.337, respectively. The lowest SSIM is on the background because the dataset is collected from surveillance videos, where the background can change drastically in different time frames. This violates our assumption that the background does not change, explaining the relatively poor performance. That said, since our goal is pose transfer, the improved performance using M-SSIM and M-LPIPS demonstrates we are more successful than even the supervised methods on Market-1501 at that task. 

\end{document}

%% file: math_commands.tex
\usepackage{amsmath,amsfonts,bm}

\def\eqref#1{equation~\ref{#1}}

\def\1{\bm{1}}

\DeclareMathAlphabet{\mathsfit}{\encodingdefault}{\sfdefault}{m}{sl}
\SetMathAlphabet{\mathsfit}{bold}{\encodingdefault}{\sfdefault}{bx}{n}

